\documentclass[letterpaper, 10 pt, conference]{ieeeconf}
\IEEEoverridecommandlockouts
\usepackage{graphicx}
\graphicspath{ {./Fig/} }
\usepackage{physics}
\usepackage{amsmath}
\usepackage{amsfonts}
\usepackage{mathtools}
\usepackage{pdfpages}
\usepackage{xurl}
\usepackage[hidelinks]{hyperref}
\usepackage{cite}
\usepackage{xcolor}
\usepackage[font=small,floatrowsep=quad,captionskip=5pt]{floatrow}
\DeclareMathSymbol{\shortminus}{\mathbin}{AMSa}{"39}

\DeclareMathOperator*{\minimize}{\text{minimize}}

\DeclareMathOperator*{\subjto}{\text{subject to}}
\usepackage{soul}
\DeclareMathAlphabet{\mathpzc}{OT1}{pzc}{m}{it}

\newcommand{\changes}[1]{#1}

\title{\textbf{Hook-Based Aerial Payload Grasping from a Moving Platform}}
\author{Péter Antal$^{1}$, Tamás Péni$^{1}$, and Roland Tóth$^{1,2}$
\thanks{This research was supported by the European Union within the framework of the National Laboratory for Autonomous Systems (RRF-2.3.1-21-2022-00002).}
\thanks{$^{1}$The authors are with the Systems and Control Lab, HUN-REN Institute for Computer Science and Control, Budapest, Hungary (email: antalpeter@sztaki.hun-ren.hu, peni@sztaki.hun-ren.hu, r.toth@tue.nl).}
\thanks{$^{2}$Roland T\'{o}th is also affiliated with the Control Systems Group of the Eindhoven University of Technology, The Netherlands.}
\vspace{-10mm}
}

\begin{document}

\maketitle

\begin{abstract}
    This paper investigates payload grasping from a moving platform using a hook-equipped aerial manipulator. First, a computationally efficient trajectory optimization based on complementarity constraints is proposed to determine the optimal grasping time. To enable application in complex, dynamically changing environments, the future motion of the payload is predicted using a physics simulator-based model. The success of payload grasping under model uncertainties and external disturbances is formally verified through a robustness analysis method based on integral quadratic constraints. The proposed algorithms are evaluated in a high-fidelity physical simulator, and in real flight experiments using a custom-designed aerial manipulator platform.
\end{abstract}

\vspace{-1mm}
\section*{Supplementary Material}
\vspace{-1mm}
A video summarizing the paper and showcasing the experiments is available at \url{https://youtu.be/PNdkaovkc5I}.
\vspace{-1mm}

\section{Introduction}
\vspace{-1mm}

There is a growing interest to perform intricate tasks with unmanned aerial vehicles (UAVs) that require robotic manipulation and direct physical interaction with the environment \cite{Ruggiero2018, Ollero2022, baraban_fruit_grasping_2021}. Payload transportation has been extensively studied in aerial manipulation, but many challenges still need to be addressed. A key challenge is to fully automate the procedure, requiring both the grasping and placement of the payload to be performed without human intervention. Furthermore, in industrial settings, the transportation has to meet strict scheduling requirements, such as being completed within a specific time window or in minimal time \cite{Maghazei2020}. 

Payload grasping from a moving platform is particularly challenging as it requires precise coordination of the manipulator with the trajectory of the payload. The problem has been first considered in \cite{spica_aerial_2012}, where the authors proposed a time-optimal planning method based on motion primitives. The applicability of the method is limited by the assumption that the trajectory of the moving platform is exactly known, and the performance of the approach has been only analyzed in simulations. In \cite{zhang_grasp_2018}, a 7 DoF robotic arm is attached to a quadcopter to grasp a moving payload. Although the experimental results show successful execution, the solution relies on baseline methods, such as PID control, therefore versatility is limited and optimality cannot be ensured. 
The current state-of-the-art work \cite{luo_time-optimal_2023} presents a time-optimal trajectory planning method for a quadcopter equipped with a 1 DoF robotic manipulator and an electromagnetic gripper. The algorithm is able to automatically find optimal handover opportunities by employing complementarity constraints. However, there is still room for improvement in certain areas, namely (i) the weight and power consumption of the manipulator reduce energy efficiency, (ii) the payload mass is not considered in the dynamic model used for planning that can degrade the performance of trajectory design, (iii) the future trajectory of the payload is assumed to be exactly known during the entire the maneuver, which is difficult to guarantee in many real-world scenarios, (iv) the trajectory optimization is complex and highly nonlinear, resulting in a computation time exceeding 10s that makes it difficult to use when real-time replanning is needed, and (v) model uncertainties and disturbances are not considered in the design, therefore robustness w.r.t. these effects is unknown.

In this work, we investigate hook-based aerial grasping from a moving platform, illustrated in Fig.~\ref{fig:intro}. The hook is attached to a rigid rod, connected to the drone via an unactuated 2 DoF joint
. Hook-based transportation has several advantages over using flexible cables \cite{Hua2021, Li2021, li_autotrans_2023, wang_impact-aware_2024} or robotic extensions \cite{zhang_grasp_2018,Thomas2014,luo_time-optimal_2023} by combining energy efficiency, good maneuverability, and autonomous grasping ability \cite{antal2024autonomous}. 

Using the hook-based manipulator, we intend to grasp a payload (with a hook attached on top) that is located on a moving unmanned ground vehicle (UGV). To address the limitations of previous works, first, instead of assuming perfect knowledge of the future motion of the payload, we use digital twin models for prediction based on the current position and velocity of the vehicle, and the task-specific reference path. Then, we explicitly consider the payload mass in trajectory planning and formulate the optimization such that it can be solved by \textit{sequential quadratic programming with real-time interations} (SQP-RTI) to enhance computational efficiency. Finally, we apply \textit{integral quadratic constraints} (IQCs) for robustness analysis of the closed-loop system governed by an LTV-LQR controller. As a result, we are able to verify successful grasping under model uncertainties and external disturbances. Our contributions are as follows:
\begin{figure}
\vspace{2mm}
    \centering \ffigbox{\caption{Payload grasping from an UGV with a hook-based manipulator.\label{fig:intro}}}{\includegraphics[width=\linewidth]{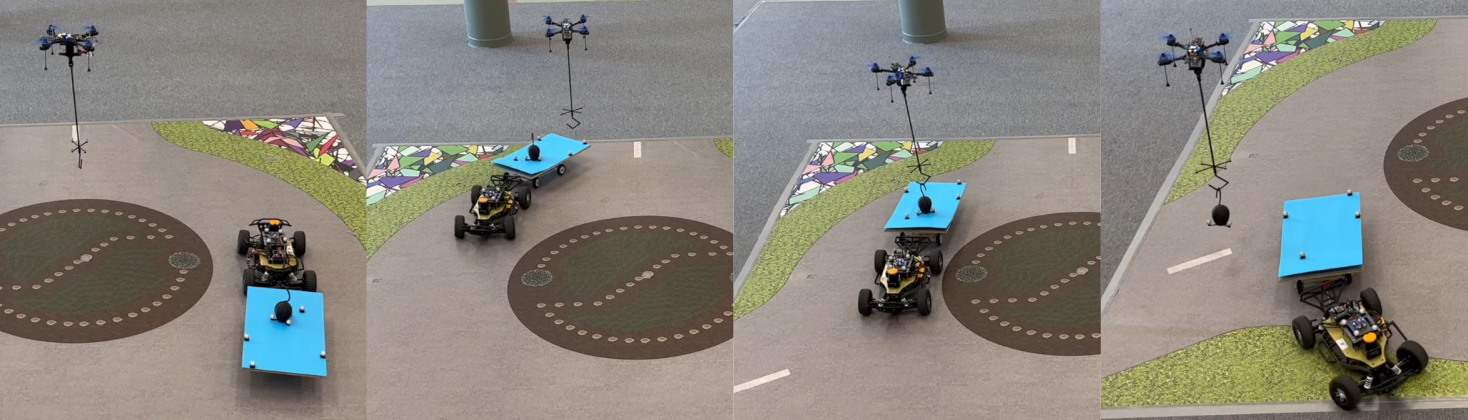}}
    \vspace{-6mm}
\end{figure}
\begin{enumerate}
    \item[C1] We propose a trajectory optimization method by complementarity constraints to automatically find the optimal grasping time instant within a pre-specified time window. The low computational cost of the SQP-RTI optimization and the digital twin-based payload motion prediction enable rapid replanning of the trajectory in response to environmental changes.
    \item[C2] We propose a robustness analysis method by integral quadratic constraints to theoretically guarantee successful grasping under uncertainties and disturbances.
    \item[C3] To demonstrate the performance of the methods, first, we conduct simulations in which the UGV navigates on an unknown, continuously changing rough terrain. Then, we perform real flight experiments.
\end{enumerate}




\vspace{-1mm}
\section{Modelling}\label{sec:model}
\vspace{-1mm}

Similar to the models commonly used to design controllers for cable-based transportation \cite{Sreenath2013, Crousaz2014}, we model the quadcopter and the payload as a rigid body and a point mass, respectively. Moreover, we neglect the mass of the rigid bar, and lump together the mass of the hook and the payload. Unlike the models used in \cite{Sreenath2013, Crousaz2014}, we take into account viscous friction in the joint connecting the hook to the drone.

\let\originalabovedisplayskip\abovedisplayskip
\let\originalbelowdisplayskip\belowdisplayskip

\begin{figure}
\vspace{2mm}
    \centering \ffigbox{\caption{Quadrotor–hook model with the three main coordinate frames.\label{fig:model}}}{\includegraphics[height=2.8cm]{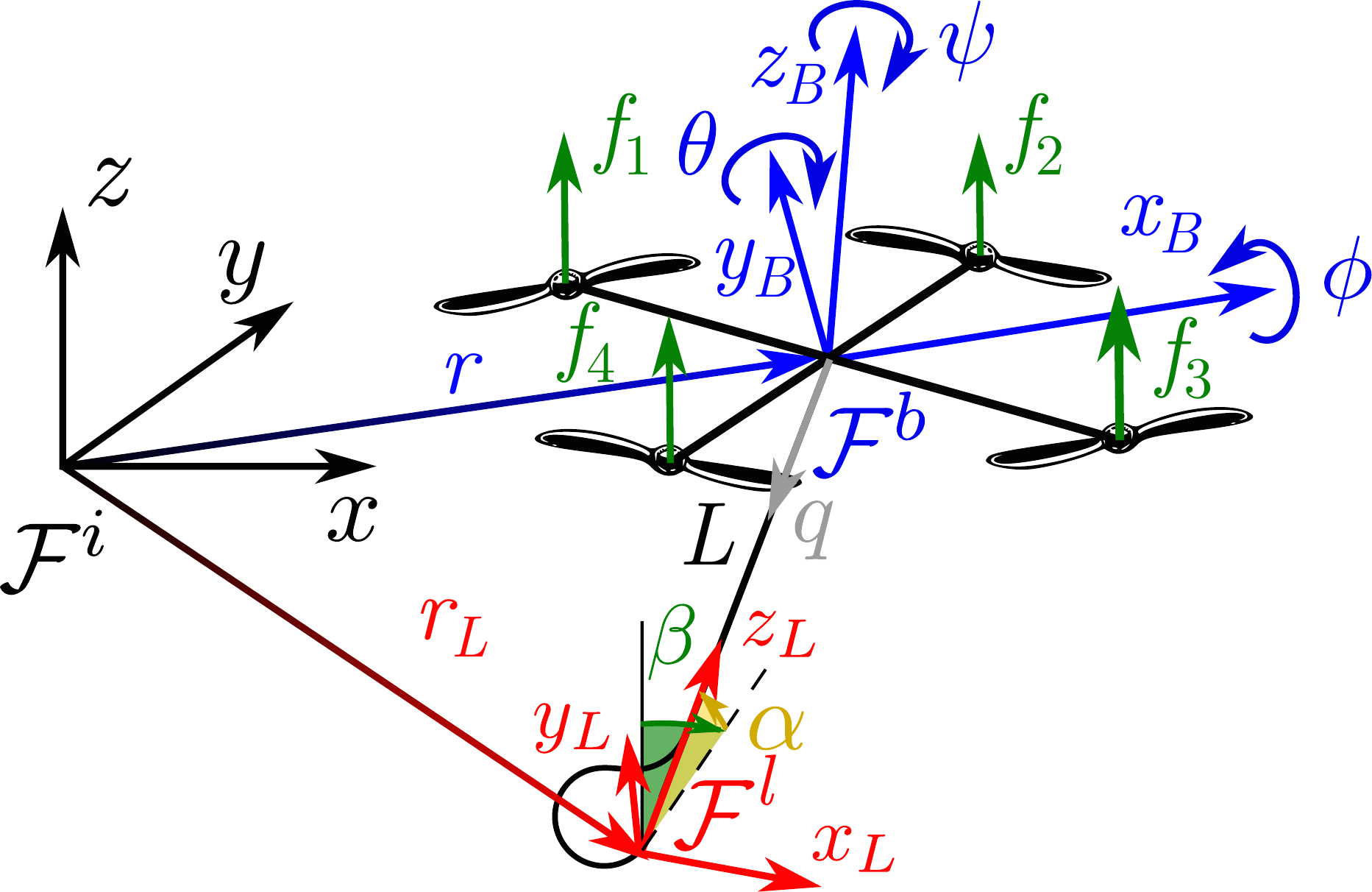}}
    \vspace{-6mm}
\end{figure}

Three main frames are used to express the system configuration: the inertial frame ($\mathcal{F}^\mathrm{i}$), the quadcopter body frame ($\mathcal{F}^\mathrm{b}$), and the payload frame ($\mathcal{F}^\mathrm{l}$), as it is illustrated in Fig.~\ref{fig:model}. We derive the model using the following state representation:
\begin{equation}\label{eq:states}
    \xi = \begin{bmatrix}
        r^\top &\!\! \dot{r}^\top &\!\! \lambda^\top &\!\! \omega^\top &\!\! \alpha &\!\! \beta & \!\!\dot\alpha &\!\! \dot\beta
    \end{bmatrix}^\top\!\!\!,\ \ \xi(t) \in \mathbb{R}^{16},
\end{equation}
where $r(t) \in \mathbb{R}^3$ is the drone position in $\mathcal{F}^\mathrm{i}$, $\lambda = [\phi \ \theta \ \psi]^\top$, $\lambda(t) \in \mathbb{R}^3$ contains the RPY (roll $\phi$, pitch $\theta$, yaw $\psi$) angles, $\omega(t) \in \mathbb{R}^3$ is the angular velocity, and $\alpha(t), \beta(t) \in \mathbb{R}$ represent the rotation of $\mathcal{F}^\mathrm{l}$ w.r.t $\mathcal{F}^\mathrm{i}$. The payload position and derivatives are expressed from the states, as follows:
\begin{subequations}\label{eq:quad_to_load_full}
    \begin{align}
    &r_\mathrm{L} = r + L q,\ \ \dot r_\mathrm{L} = \dot r + L \dot q, \ \ \ddot r_\mathrm{L} = \ddot r + L \ddot q,\label{eq:quad_to_load}\\
    &q = \begin{bmatrix} -\cos\alpha\, \sin \beta & \sin\alpha & -\cos\alpha \,\cos\beta  \end{bmatrix}^\top,\label{eq:q_al_bet}
\end{align}
\end{subequations}
where $q(t)\in\mathbb{R}^3$ is the unit vector pointing from the quadcopter CoM to the payload CoM. 

To derive the dynamic model, we take the equations of motion from \cite{sreenath_trajectory_2013} corresponding to the payload suspension:
\begin{equation}\label{eq:dyn}
    \begin{split}
        & m \ddot r = F R e_3 - m g e_3 + Tq,\\
         m_\mathrm{L} \ddot r_\mathrm{L} = &-T q - m_\mathrm{L} g e_3,\ 
        J \dot{\omega}+\omega \times J \omega=\tau,
    \end{split}
\end{equation}
where $m, m_\mathrm{L}$ are the mass of the quadcopter and the payload (with the hook), $e_3 = [\ 0 \ 0 \ 1 \ ]^\top$, $T(t) \in \mathbb{R}$ is the force on the rigid bar, $R(t) \in \mathrm{SO}(3)$ is the rotation matrix on the special orthogonal group, $J = \mathrm{diag}(J_\mathrm{x}, J_\mathrm{y}, J_\mathrm{z})$ is the inertia matrix of the quadcopter, assumed to be diagonal, and $F(t) \in \mathbb{R}, \tau(t) \in \mathbb{R}^3$ are force and torque inputs generated by the propellers. Based on \cite{Sreenath2013}, the dynamics of $q$ are described, as follows:
\begin{equation}\label{eq:ddot_q}
    \ddot{q} = (m L)^{-1} \left( q \times (q\times F R e_3) \right) - (\dot{q}^\top \dot{q})q.
\end{equation}
The position dynamics are formulated by combining \eqref{eq:quad_to_load_full}-\eqref{eq:ddot_q}:
\begin{equation}\label{eq:ddot_r}
\begin{split}
        \ddot r =& f_\mathrm{r}(q, \dot q, R, F) =   (m + m_\mathrm{L})^{-1} F R e_3 - g e_3 \\
        &- \tfrac{m_\mathrm{L} L}{m + m_\mathrm{L}} \left( \tfrac{1}{m L} \left( q \times (q\times F R e_3) \right) - (\dot{q}^\top \dot{q})q \right).
\end{split}
\end{equation}
In the next step, we formulate the equations of the two revolute joints by expressing $\alpha$ and $\beta$ from \eqref{eq:q_al_bet}, differentiating it two times, and substituting \eqref{eq:ddot_q} into $\ddot q$, as follows: $[\  \ddot\alpha \ \ddot\beta \ ]^\top = f_\mathrm{L} (q, \dot q, R, F)$. Due to the use of inverse trigonometric functions, $f_\mathrm{L}$ is singular if the orientation of the rod becomes horizontal. \changes{However, 
the considered grasping and transportation tasks does not require such large swing angles, therefore this effect does not limit the performance and applicability of the method.}

We complete the model by putting together the above equations and adding a viscous friction term to the dynamics of the revolute joints:
\begin{equation}\label{eq:nl_gen}
        \dot\xi = f(\xi, u) = \begin{bmatrix}
        \dot r \\ 
        f_\mathrm{r}(q, \dot q, R, F)\\
        Q^{-1} \omega \\
        J^{-1}(\tau - \omega \times J \omega)\\
        [\ \dot\alpha \ \dot\beta\ ]^\top \\
        f_\mathrm{L} (q, \dot q, R, F) - b [\ \dot \alpha\ \dot\beta\ ]^\top
    \end{bmatrix},
\end{equation}
where \changes{$u =  [\ F \ \tau^\top ]^\top$ is the control input}, $Q(t) \in \mathbb{R}^{3\times 3}$ is the standard transformation matrix from the derivatives of the Euler angles to the angular velocity \cite{Beard2012}, and $b\in\mathbb{R}^+$ is the viscous friction coefficient, the numerical value of which is obtained by measuring the logarithmic decrement of the hook oscillations while the quadcopter body is fixed \cite{thomson_vibration_2018}.

Although \eqref{eq:nl_gen} captures the full dynamics of the aerial manipulator, the position and velocity of the payload are not explicitly expressed by the state vector. In trajectory planning, as multiple constraints are required for the payload states, we modify the state vector as follows:
\begin{equation}\label{eq:states_traj}
    \tilde{\xi} = \begin{bmatrix}
        r_\mathrm{L}^\top &\!\! \dot{r}_\mathrm{L}^\top &\!\! \lambda^\top &\!\! \omega^\top &\!\! \alpha &\!\! \beta &\!\! \dot\alpha &\!\! \dot\beta
    \end{bmatrix}^\top\!\!,\ \ \tilde{\xi}(t) \in \mathbb{R}^{16}.
\end{equation}
Then, if \eqref{eq:ddot_q}, \eqref{eq:ddot_r} are substituted into \eqref{eq:quad_to_load}, we can arrive at
\begin{align*}
    \ddot r_\mathrm{L}\! =\! f_{\mathrm{r_L}}(q, \dot q, R, F) \! =\! \tfrac{1}{m\! +\! m_\mathrm{L}} (q^\top F R e_3 - m L \dot{q}^\top \dot{q}) q - g e_3.
\end{align*}
Finally, the complete equations of motion are as follows:
\begin{equation}\label{eq:nl_gen_traj}
        \dot{\tilde{\xi}} = \tilde f(\tilde{\xi}, u) ,
\end{equation}
where $\tilde f$ is the same as $f$ in \eqref{eq:nl_gen}, except for replacing $\dot r$ by $\dot r_\mathrm{L}$ and $f_\mathrm{r}$ by $f_\mathrm{r_L}$.
\vspace{-1mm}
\section{Trajectory Planning}\label{sec:planning}
\vspace{-1mm}
\subsection{Payload motion prediction}
\vspace{-1mm}

In prior works, the future trajectory of the moving payload is typically assumed to be exactly known \cite{spica_aerial_2012, luo_time-optimal_2023, zhang_grasp_2018}. In contrast, we only assume that the task-specific reference path of the UGV and the structure of the environment are known 
before grasping, \changes{with enough time to replan the trajectory}. For prediction, we use the digital twin model of the UGV, the manipulator, and the payload, which has been implemented in MuJoCo \cite{todorov2012mujoco}, a high-fidelity physics engine that is able to simulate complex dynamics with collisions and contacts. 

A key challenge in digital twin-based prediction is the accurate initialization of the simulation. Although we assume that all states can be measured or computed from the kinematics, imperfect measurements may cause significant variations in the simulated contact forces. To enhance robustness w.r.t. initial conditions, we simulate the UGV dynamics in closed loop with a low-level controller. In simulation experiments, we test this robustness by adding white Gaussian noise to the initial states. Our results, detailed in Section~\ref{sec:simu}, show that the grasping is accomplished in dynamically changing environments that confirms practical effectiveness of the approach. Nonetheless, more advanced algorithms could be employed for initialization, see e.g. \cite{lowrey_physically-consistent_2014}.


\vspace{-0.5mm}
\subsection{Planning concept}
\vspace{-0.5mm}
\let\originalabovedisplayskip\abovedisplayskip
\let\originalbelowdisplayskip\belowdisplayskip

\setlength{\abovedisplayskip}{1ex plus 1pt minus 1pt}
\setlength{\belowdisplayskip}{\abovedisplayskip}


Motivated by the strict schedule of industrial processes, we define a time window, within which the quadcopter has to grasp the payload:
\begin{equation}
    0 < \underline{T}_\mathrm{g} \leq T_\mathrm{g} \leq \overline{T}_\mathrm{g} < T_\mathrm{f},
\end{equation}
where the beginning and end of the time window ($\underline{T}_\mathrm{g}, \overline{T}_\mathrm{g}$), together with the duration ($T_\mathrm{f}$) are fixed in the problem description, and the grasp time ($T_\mathrm{g}$) is a decision variable. 

\begin{figure}
\vspace{2mm}
    \centering \ffigbox{\caption{Left: 3 phases of the motion trajectory, right: grasping conditions.
    \label{fig:planning}}}{\includegraphics[width=.5\linewidth]{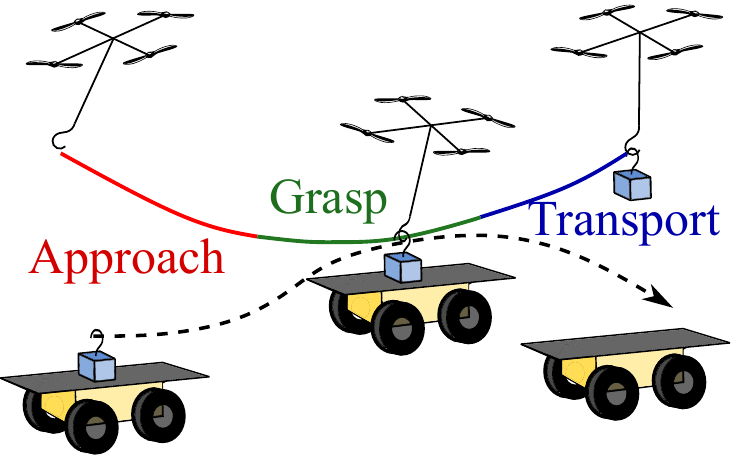}\hspace{3mm}
    \includegraphics[width=.26\linewidth]{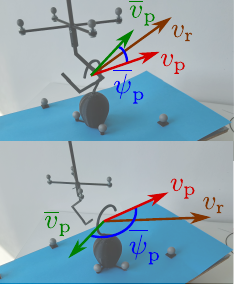}}
    \vspace{-6mm}
\end{figure}

For payload grasping and transportation, we propose a nonlinear optimization-based trajectory planning method. First, as it is illustrated on Fig.~\ref{fig:planning}, we divide the motion to three phases: 1) approach the moving payload; 2) attach the hook; 3) transport the payload above the target location. Phase 1 takes until the beginning of the grasping time window. Meanwhile, the quadcopter has to approach the payload to get ready for grasping. In Phase 2, the aerial manipulator has to grasp the payload by attaching the hook. 
At the end of the grasping time window, Phase 3 starts, where the goal is to transport the payload above the target location.

To perform successful grasping, the motion of the quadcopter is coordinated with the trajectory of the payload. First, the hook of the quadcopter and the payload are aligned to be perpendicular at the grasping time instant. Second, at that point, a target velocity is prescribed for the hook, defined as the sum of the payload velocity and a relative velocity term that is parallel to the body $x$ axis of the payload. These conditions are formulated as follows:
\begin{align}\label{eq:grasp_cond}
    [\psi_\mathrm{r}, v_\mathrm{r}]\!  =\!  \begin{cases}
        [\psi_\mathrm{p}, (1\! +\! \zeta R_\mathrm{z}(\overline \psi_\mathrm{p})) v_\mathrm{p}] &\text{if $|\overline \psi_\mathrm{p}| \leq \frac{\pi}{2}$}\\
        [\psi_\mathrm{p}\!\!  \shortminus  \vartheta,  (1\! +\! \zeta R_\mathrm{z}(\overline \psi_\mathrm{p}\!\! \shortminus \vartheta)) v_\mathrm{p}]\! &\text{otherwise}
    \end{cases}
\end{align}
where $\psi_\mathrm{p}(t)\in\mathbb{R}, v_\mathrm{p}(t)\in\mathbb{R}^3$ are the yaw angle and velocity of the payload, $\overline \psi_\mathrm{p}(t)\in\mathbb{R}$ is the angle between the velocity and body $x$ axis of the payload, $\overline v_\mathrm{p}(t)\in\mathbb{R}^3$ denotes $v_\mathrm{p}$ rotated by $\overline \psi_\mathrm{p}$ around the $z$ axis of the inertial frame, $\psi_\mathrm{r}(t)\in\mathbb{R}, v_\mathrm{r}(t)\in\mathbb{R}^3$ are the target yaw and velocity of the hook, $\vartheta = \mathrm{sgn}(\overline \psi_\mathrm{p})\pi/2$, and $R_\mathrm{z}$ denotes the rotation matrix around the $z$ axis of the inertial frame. The user-defined $\zeta \in \mathbb{R}^+$ determines the relative velocity of the hook, increasing its value leads to more aggressive grasping. The grasping conditions are illustrated in Fig.~\ref{fig:planning}.

\setlength{\abovedisplayskip}{\originalabovedisplayskip}
\setlength{\belowdisplayskip}{\originalbelowdisplayskip}

\vspace{-0.5mm}
\subsection{Trajectory optimization}
\vspace{-0.5mm}

\textbf{Discretization:}  We formulate the trajectory optimization in discrete time to exploit the benefits of multiple shooting methods \cite{bock_multiple_1984}. The dynamic model \eqref{eq:nl_gen_traj} is discretized by 4$^\mathrm{th}$ order Runge-Kutta method with sampling time $h \in \mathbb{R}^+$ as $\tilde\xi_\mathrm{d}(k+1) = \tilde f_{\mathrm{RK4}} (\tilde\xi_\mathrm{d}(k), u_\mathrm{d}(k))$, where $k\in \mathbb{Z}$ is the discrete time index with $t\!=\!kh$, and $\tilde\xi_\mathrm{d}(k) \!\in\! \mathbb{R}^{16}, u_\mathrm{d}(k) \!\in\! \mathbb{R}^{4}$ are the states and inputs. The time window parameters are also adjusted to the sampling time as $\underline{N}_\mathrm{g} = [\underline{T}_\mathrm{g}/h]$, $\overline{N}_\mathrm{g} \!=\! [\overline{T}_\mathrm{g}/h]$, $N_\mathrm{f}\!=\! [T_\mathrm{f}/h]$, where $[x]$ is rounding to the nearest integer.

\textbf{Cost function:} For each of the three phases, we introduce different cost terms and constraints, according to the current planning objective. Then, we combine the three phases into one optimization problem. We use quadratic running costs:
\begin{equation}\label{eq:running_cost}
   l_i(k) =  \| \tilde\xi_\mathrm{d}(k)- \tilde\xi_\mathrm{r}^i(k) \|_{W_{\mathrm{Q}}^i}^2 +\| u_\mathrm{d}(k) - u_\mathrm{r}(k) \|_{W_{\mathrm{R}}^i}^2
\end{equation}
for $i\!=\!1,2,3$, where $W_{\mathrm{Q}}^i \in \mathbb{R}^{16 \times 16}, W_{\mathrm{R}}^i \in \mathbb{R}^{4 \times 4}$ are positive semidefinite weight matrices and $\tilde\xi_\mathrm{r}^i, u_\mathrm{r}$ denote target states and inputs. The input term is used for regularization w.r.t. to hovering, therefore we set $u_\mathrm{r}(k) = [\ (m + m_\mathrm{L}) g \ 0 \ 0 \ 0\ ]^\top$. Next, the target states are defined for each phase.

The objective of Phase 1 is to approach the moving payload, which we solve by following the trajectory of the ground vehicle with a delay of $T_\mathrm{d} \in \mathbb{Z}^{+}$ time steps. Accordingly, we define the target state as follows:
\begin{align*}
    \tilde\xi_\mathrm{r}^1(k)\! =\! [\  r_\mathrm{p}(k - T_\mathrm{d})^\top \ \ 0_{5} \ \ \psi_\mathrm{r} (k - T_\mathrm{d}) \ \ 0_{8}\ ]^\top\!\!\!\!,\ \ k\in [0, \underline{N}_\mathrm{g}]
\end{align*}
where $r_\mathrm{p}$ is the payload position and $\psi_\mathrm{r}$ is given by \eqref{eq:grasp_cond}. 

In Phase 2, the payload needs to be grasped from the moving platform. Consequently, we set the target state based on \eqref{eq:grasp_cond}, as follows:
\begin{align*}
    \tilde\xi_\mathrm{r}^2(k) = [\ r_\mathrm{p}(k)^\top \ \ v_\mathrm{r}(k)^\top \ \ 0_{2} \ \ \psi_\mathrm{r} (k) \ \ 0_{8}\ ]^\top\!\!\!\!,\ \ k\in [\underline{N}_\mathrm{g}, \overline{N}_\mathrm{g}]
\end{align*}

In Phase 3, the objective is to transport the payload to the user-defined target location $r_\mathrm{L, f}$ with target yaw $\psi_\mathrm{f}$, therefore we define $\tilde\xi_\mathrm{r}^3(k) = \tilde\xi_\mathrm{d, f} = [\ r_\mathrm{L, f}^\top\ 0_5\ \psi_\mathrm{f}\ 0_8\ ]^\top$ for $k\in[\overline{N}_\mathrm{g}, {N}_\mathrm{f}]$.

\setlength{\abovedisplayskip}{0.7ex plus 1pt minus 1pt}
\setlength{\belowdisplayskip}{\abovedisplayskip}
\textbf{Constraints:} We bound the states and control inputs of the manipulator, as follows:
\begin{align}
    u_\mathrm{d, min} \leq u_\mathrm{d}(k) \leq u_\mathrm{d, max},\ \tilde\xi_\mathrm{d, min} \leq  \tilde\xi_\mathrm{d}(k) \leq  \tilde\xi_\mathrm{d, max},
\end{align}
where $u_\mathrm{d, min}, u_\mathrm{d, max}$ represent the actuator limits of the quadcopter, and $\tilde\xi_\mathrm{d, min}, \tilde\xi_\mathrm{d, max}$ are specified by the user based on e.g. the available flying space or safety considerations.

To ensure that the hook of the quadcopter and the payload get attached, we employ \emph{complementarity constraints}, similar to \cite{luo_time-optimal_2023}. For this, we introduce the index variable $\varepsilon(k) \in \mathbb{R}$ to indicate whether the grasping conditions given by \eqref{eq:grasp_cond} are satisfied. Then, we formulate the following constraints:
\begin{align}\label{eq:compl}
&\varepsilon(k) f_\mathrm{gr}(k) = 0, \quad \varepsilon(k) \in [0, 1]\\
&f_\mathrm{gr}\!(k) \!=\! \|  r_\mathrm{L}\!(k) \shortminus r_\mathrm{p}\!(k) \| + \| \psi(k) \shortminus \psi_\mathrm{r}\!(k) \| 
+ \| \dot r_\mathrm{L}\!(k) \shortminus v_\mathrm{r}\!(k) \|.\nonumber
\end{align}
To ensure that $\varepsilon(k)\! >\! 0$ and $f_\mathrm{gr}(k)\! =\! 0$ at least once in Phase~2, we introduce the cumulative index variable $\kappa(k)\! \in\! \mathbb{R}$ with corresponding dynamics, as follows:
\begin{align}\label{eq:kappa}
    \kappa (k+1) = \kappa(k) - \varepsilon(k),\quad \kappa(0) = 1, \quad \kappa (N_\mathrm{f}) = 0.
\end{align}
Since $\kappa(0) > \kappa (N_\mathrm{f})$, $\varepsilon(k)$ is forced to be nonzero at some point of the trajectory, implying that the grasping conditions are satisfied.\footnote{The constraint formulation would allow for $\varepsilon(k)>0$ at multiple times, enabling multiple grasp points. However, we restrict $[\underline{T}_\mathrm{g}, \overline{T}_\mathrm{g}]$ to ensure only one dynamically feasible grasp point, where $\varepsilon(k)=1$.} 
In practice, strict complementarity constraints are difficult to handle by numerical solvers \cite{luo_time-optimal_2023}, therefore we relax \eqref{eq:compl} by introducing a slack variable $\nu$, as follows:
\begin{align}\label{eq:compl_rel}
    \varepsilon(k) ( f_\mathrm{gr}(k) - \nu(k) ) = 0, \quad 0 \leq \nu(k) \leq \nu_\mathrm{max},
\end{align}
where $\nu_\mathrm{max} \in \mathbb{R}$ is set to a small constant, e.g. $10^{-3}$.

\textbf{Optimal control problem: } We formulate trajectory planning as a finite-horizon optimal control problem (OCP) by putting together the cost function and constraints. For this, we incorporate the complementarity variables into the dynamic model, resulting in the augmented states $\bar\xi_\mathrm{d} = [\ \tilde\xi_\mathrm{d}^\top \ \kappa\ ]^\top$, augmented inputs $\bar u_\mathrm{d} = [\ u_\mathrm{d}^\top \ \varepsilon \ \nu \ ]^\top$, and augmented dynamics $\bar f(\bar\xi_\mathrm{d}, \bar u_\mathrm{d}) = [\ \tilde f_{\mathrm{RK4}}(\bar\xi_\mathrm{d}, \bar u_\mathrm{d})^\top \ \kappa-\varepsilon \ ]^\top$. Now, \eqref{eq:running_cost}-\eqref{eq:compl_rel} are combined to formulate the OCP:
\begin{subequations}\label{eq:planning_ocp}
    \begin{align} 
    \minimize_{\substack{\bar \xi_\mathrm{d}(k), \bar u_\mathrm{d}(k), N_\mathrm{g} \\k = 0, \dots, N_\mathrm{f}-1}} & J = \!\! \sum_{k=0}^{\underline{N}_\mathrm{g}\!-\!1} l_1(k) + \!\! \sum_{k=\underline{N}_\mathrm{g}}^{\overline{N}_\mathrm{g}\!-\!1}\! l_2(k) + \!\! \sum_{k=\overline{N}_\mathrm{g}}^{N_\mathrm{f}\!-\!1}\! l_3(k)  \\
    \subjto \;\; & \bar \xi_\mathrm{d}(0) = \bar \xi_\mathrm{d, 0},\ \bar \xi_\mathrm{d}(N_\mathrm{f}) = \bar \xi_\mathrm{d, f}\\
    & \varepsilon(k) ( f_\mathrm{gr}(k)\! -\! \nu(k) )\! =\! 0 \ \ k\in [\underline{N}_\mathrm{g} ,\overline{N}_\mathrm{g}]\\
    & \varepsilon(k) = 0 \text{\phantom{00000000000i,,}} \ k\notin [\underline{N}_\mathrm{g} ,\overline{N}_\mathrm{g}]\\
    & \bar \xi_\mathrm{d}(k+1) = \bar f (\bar \xi_\mathrm{d}(k), \bar u_\mathrm{d}(k))\\
    & m_\mathrm{L} = m_\mathrm{hook} + (1-\kappa) m_\mathrm{load}\label{eq:planning_ocp_ml}\\
    & \bar \xi_\mathrm{d, min} \leq \bar \xi_\mathrm{d}(k) \leq \bar \xi_\mathrm{d, max}\\
    & \bar u_\mathrm{d, min} \leq \bar u_\mathrm{d}(k) \leq \bar u_\mathrm{d, max}
\end{align}
\end{subequations}
where $\bar \xi_\mathrm{d, 0}, \bar \xi_\mathrm{d, f}, \bar \xi_\mathrm{d, min}, \bar \xi_\mathrm{d, max}, \bar u_\mathrm{d, min}, \bar u_\mathrm{d, max}$ correspond to the boundary conditions given by \eqref{eq:compl}-\eqref{eq:compl_rel}. Note that \eqref{eq:planning_ocp_ml} ensures that the mass of the load is added to the mass of the hook automatically after the grasping time instant, without knowing the value of $N_\mathrm{g}$ a priori. Hence, we make sure that the effect of the payload on the dynamics is directly considered in the physical model, therefore the optimized trajectory is dynamically feasible (in contrast e.g. to \cite{luo_time-optimal_2023}).

To solve Optimization~\eqref{eq:planning_ocp}, we use the \textit{sequential quadratic programming with real-time iterations} (SQP-RTI) solver of \texttt{acados} \cite{verschueren_acadosmodular_2022}. Although the dynamics and complementarity constraints are nonlinear, the cost function is quadratic and most constraints are linear. This leads to a computation time of around 0.4 s on a standard laptop, which enables real-time replanning. To replan the trajectory at time step $k=N_\mathrm{rp}$, we shift the time window by subtracting $N_\mathrm{rp}$ from $\underline{N}_\mathrm{g} ,\overline{N}_\mathrm{g},N_\mathrm{f}$, and solve \eqref{eq:planning_ocp} on a shrinked horizon \cite{greer_shrinking_2020}.

\setlength{\abovedisplayskip}{\originalabovedisplayskip}
\setlength{\belowdisplayskip}{\originalbelowdisplayskip}

\vspace{-1mm}
\section{Trajectory Tracking}\label{sec:control}
\vspace{-1mm}

Due to its generality, low complexity, and computational efficiency, we design a \textit{linear time-varying linear-quadratic regulator} (LTV-LQR) \cite{Kwakernaak} to follow the 
reference trajectory. To synthesize the controller, we linearize the equations of motion \eqref{eq:nl_gen} along the reference by first-order Taylor expansion:
 $\dot\xi \!\approx\! f(\xi^*\!, u^*) + A (\xi \!-\! \xi^*) + B (u \!-\! u^*),$ 
where $A(t) \in \mathbb{R}^{16\times 16}, B(t) \in \mathbb{R}^{16\times 4}$ are the time-varying state and input matrices, respectively. The reference states and inputs $\xi^*(t), u^*(t)$ are obtained by solving Optimization~\eqref{eq:planning_ocp}, interpolating the discrete-time states and inputs by splines, and applying \eqref{eq:quad_to_load} to transform between the state representations \eqref{eq:states_traj} and \eqref{eq:states}. With error state $\eta = \xi - \xi^*$ and input $v = u - u^*$, the LTV error dynamics are written, as follows:
\begin{align}\label{eq:lin_dyn}
    \dot \eta(t) = A(t) \eta(t) + B(t) v(t).
\end{align}


Then, the optimal feedback gain matrix is computed as the solution of a finite-horizon LQ optimization problem:
\begin{align}\label{eq:ltv_lq_opt}
    \minimize_{v} \quad & \int_{0}^{T_\mathrm{f}} \left(\| \eta(t) \|_{\mathcal W_\mathrm{Q}}^2 + \| v(t)\|_{\mathcal W_\mathrm{R}}^2 \right) \mathrm{d} t + \| \eta(T_\mathrm{f}) \|_{\mathcal W_\mathrm{F}}^2\nonumber\\
    \subjto \quad & v(t) = - K(t) \eta(t), \ \text{\eqref{eq:lin_dyn}}
\end{align}
where $\mathcal W_{\mathrm{Q}}, \mathcal W_{\mathrm{F}} \in \mathbb{R}^{16 \times 16}, \mathcal W_{\mathrm{R}} \in \mathbb{R}^{4 \times 4}$ are positive semidefinite state and input weight matrices and $K(t) \in \mathbb{R}^{4 \times 16}$ is the time-varying feedback gain matrix. 


\begin{figure}
\vspace{2mm}
    \centering \ffigbox{\caption{Left: Interconnection of nominal LTV system $G$ and uncertainty $\Delta$ with disturbance input $d$ and performance output $e$. Right: Extended LTV system with IQC filter $\Psi$. \cite{seiler_finite_2019}
    \label{fig:unc_sys}}}{\includegraphics[width=0.72\linewidth]{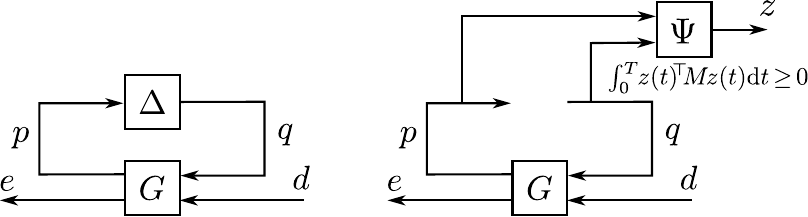}}
    \vspace{-6mm}
\end{figure}

\vspace{-1.5mm}
\section{IQC-based Robustness Analysis}\label{sec:anal}
\vspace{-1.5mm}

\subsection{Uncertain dynamics}
\vspace{-1.5mm}

The equations of motion given by \eqref{eq:nl_gen} describe the nominal dynamics of the aerial manipulator. However, in reality, the exact value of the parameters are hard to determine, and other uncertainties may also be present, e.g. due to unmodeled actuator dynamics or aerodynamics \cite{torrente_data-driven_2021}. These uncertainties may deteriorate the precision of the motion which could eventually lead to unsuccessful payload grasping. In this section, we introduce an uncertain model of the system and compute its \emph{robust $\mathcal{L}_2$-to-Euclidean gain} \cite{seiler_finite_2019} to characterize the worst-case hook position error at the grasping time instant. If the worst-case error is smaller than the radius of the hook, then the grasping is guaranteed to be successful.


We consider the $x, y$ inertia elements and the viscous friction of the joint as uncertain parameters collected to $\tilde{\delta}=[\ J_\mathrm{x}\ J_\mathrm{y}\  b\ ]^\top$ with nominal values $\varphi_\mathrm{0}=[\ J_\mathrm{x, 0}\ J_\mathrm{y, 0}\ b_\mathrm{0}\ ]^\top$ and uncertainty range $\bar\varphi\in\mathbb{R}^3$ such that $\tilde{\delta} \in [\varphi_\mathrm{0}-\bar\varphi,\varphi_\mathrm{0}+\bar\varphi]$. We normalize the uncertainties as $\tilde{\delta} = \varphi_0 + \mathrm{diag}(\bar \varphi) \delta$. Then, we take the closed-loop nominal LTV system $G$ given by \eqref{eq:lin_dyn}, \eqref{eq:ltv_lq_opt} and formulate a \emph{linear fractional representation} 
by lifting the uncertainties to $\Delta = \mathrm{diag}(\delta_1 I_{n_1}, \delta_2 I_{n_2}, \delta_3 I_{n_3})$, where $n_1\!=\!n_2\!=\!3, n_3\!=\!2$ are the corresponding signal dimensions. To represent unmodeled aerodynamics, we consider a scalar force disturbance $d(t)\! =\! \delta F (t), \| d \|_{2, [0, T]}\! \leq\! \beta$ added to the thrust 
input of the quadrotor, where $\| \cdot \|_{2, [0, T]}$ is the finite-horizon $\mathcal{L}_2[0, T]$ norm of a signal \cite{seiler_finite_2019}. 
According to the objective of this analysis, the output $e(t)$ is the position error of the hook. The block diagram of the resulting uncertain system is shown in Fig.~\ref{fig:unc_sys}, and it is described by the finite-time state-space model $G_\Delta$: 
\begin{align}
\begin{split}\label{eq:unc_ltv_dyn}
    \dot \eta(t) &= A_\mathrm{G}(t) \eta(t) + B_\mathrm{G1}(t) q(t) + B_\mathrm{G2}(t) d(t),\\
    p(t) &= C_\mathrm{G1}(t) \eta(t) + D_\mathrm{G11}(t) q(t) + D_\mathrm{G12}(t) d(t),\\
    e(t) &= C_\mathrm{G2}(t) \eta(t) + D_\mathrm{G21}(t) q(t) + D_\mathrm{G22}(t) d(t)   
\end{split}
\end{align}
for $t \in [0, T]$, where $A_\mathrm{G}(t) = A(t) - B(t) K(t)$, and $p(t)\in\mathbb{R}^{n_\mathrm{p}}, q(t)\in\mathbb{R}^{n_\mathrm{q}}$ are the latent variables corresponding to $\Delta$ with $n_\mathrm{p}=n_\mathrm{q}=n_1+n_2+n_3$. 

\begin{figure}
   \CenterFloatBoxes
    \begin{floatrow}
        \ffigbox[\FBwidth]{\caption{UGV reference path (blue), and initial drone positions for robustness analysis (orange).\label{fig:paperclip}\vspace{-4mm}}}{%
            \includegraphics[width=3.7cm]{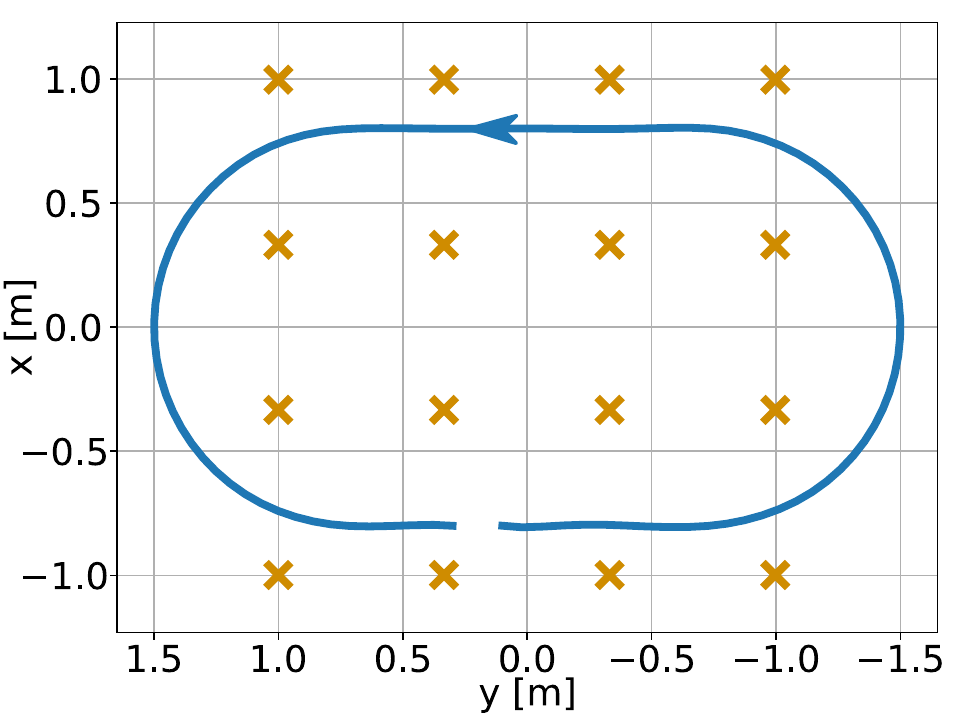}
        }
        \killfloatstyle\ttabbox[\FBwidth]{%
            \caption{Physical parameters of the \\ \vspace{-1mm} manipulator.\label{tab:drone_param} \vspace{-2mm}}
        }{\scriptsize \begin{tabular}{lrl}
                \hline
                $m$ & $6.05\cdot 10^{-1}$ & kg \\ 
                $m_\mathrm{hook}$ & $1\cdot 10^{-2}$ & kg \\ 
                $J_\mathrm{x,0}$  &  $1.5\cdot 10^{-3}$ & kg m$^2$ \\ 
                $\delta J_\mathrm{x}$ & $3\cdot 10^{-4}$ & kg m$^2$ \\ 
                $J_\mathrm{y,0}$ &  $1.45\cdot 10^{-3}$ & kg m$^2$ \\ 
                $\delta J_\mathrm{y}$  &$3\cdot 10^{-4}$ & kg m$^2$  \\ 
                $J_\mathrm{z}$ &  $2.66\cdot 10^{-3}$ & kg m$^2$ \\ 
                $L$  & $4\cdot 10^{-1}$ & m \\ 
                $b_0$  &  $2.9\cdot 10^{-1}$& rad s \\ 
                $\delta b$  & $1\cdot 10^{-1}$& rad s\\
                \hline
            \end{tabular}}
    \end{floatrow}
    \vspace{-6mm}
\end{figure}

\vspace{-1mm}
\subsection{Integral quadratic constraints}
\vspace{-1mm}

We characterize the performance of the uncertain LTV system $G_\Delta$ using the finite-horizon \textit{$\mathcal{L}_2$-to-Euclidean gain}, defined as $\| G_\Delta \|_{E,[0, T]} = \sup_{\|d\|_{2,[0,T]}\neq 0}\left\{\frac{\|e(T)\|}{\|d\|_{2,[0,T]}}  \right\}$, where $\|\cdot\|$ is the 2-vector norm. To compute the performance metric, we employ \emph{integral quadratic constraints} (IQCs) \cite{megretski_system_1997}. 

Let $\Psi \in \mathbb{RH}_\infty^{n_\mathrm{z}\times(n_\mathrm{p} + n_\mathrm{q})}$ be a stable rational proper transfer function with $z = \Psi(p,q)$ and $M: [0, T]\rightarrow \mathbb{S}^{n_\mathrm{z}}$ a piecewise continuous matrix function. The input-output signals $p\in  \mathcal{L}_2^{n_\mathrm{p}}[0, T]$ and $q = \Delta(p)$ 
satisfy the time-domain IQC defined by $(\Psi, M)$ %
if it holds for all $\Delta : \mathcal{L}_2^{n_\mathrm{p}}[0, T] \rightarrow \mathcal{L}_2^{n_\mathrm{q}}[0, T]$ that
\begin{align}\label{eq:iqc}
    \int_0^T z(t)^\top M(t) z(t) \mathrm{d} t \geq 0,
\end{align}
where $p, q$ are filtered through $\Psi$ with zero initial conditions. Our goal is to represent the uncertain relation by finding $M(t)$ such that \eqref{eq:iqc} is satisfied for any $z(t)$ produced by $\Delta$ with the smallest conservativism. 
As our system contains parametric uncertainty, we search for $M$ with a block diagonal structure $M = \mathrm{diag}(M^1, M^2, M^3)$, where each block is parametrized as follows \cite{megretski_system_1997, veenman_robust_2016}:
\begin{align}\label{eq:iqc_param}
    M^i \coloneqq \begin{bmatrix}
        M_{11}^i & M_{12}^i \\ (M_{12}^i)^\top & -M_{11}^i 
    \end{bmatrix} \quad \text{for } i=1,2,3
\end{align}
where $M_{11}^i\!\! =\! (M_{11}^i)^\top \!\succeq\! 0$ and $M_{12}^i \!=\! -(M_{12}^i)^\top$ are constant real matrices. Inspired by \cite{seiler_finite_2019}, we parametrize $\Psi$ as $\Psi = \mathrm{diag}(\Psi^1, \Psi^2, \Psi^3),$ $\Psi^i = \mathrm{diag}(\psi_{11}^i, \psi_{11}^i),$ $\psi_{11}^i = [\ 1 \ \frac{1}{s+\lambda}\ ]^\top,$ $\lambda>0$ for $i=1, 2, 3$. 
Then, during the analysis, we fix $\Psi$ by selecting the value of $\lambda$ manually, and optimize over $M$.



\vspace{-1.5mm}

\subsection{Robustness analysis}
\vspace{-1mm}

Our goal is to find the radius $\rho\in\mathbb{R}^+$ of the smallest ball that covers all possible outputs at time $T$ $(e(T))$ given that the initial states lie within an ellipsoidal set $\mathcal{E}_0 = \{ \eta \in \mathbb{R}^{n_\eta}: \eta^\top A_0 \eta \leq 1, A_0 = A_0^\top \succ 0\}$ and the dynamics satisfy \eqref{eq:unc_ltv_dyn}-\eqref{eq:iqc}. To solve this problem, an efficient and scalable algorithm is proposed in \cite[Ch. 3.1]{moore_finite_2015} by computing the robust $\mathcal{L}_2$-to-Euclidean gain. First, a time-varying quadratic storage function is defined. Then, a dissipation inequality is formulated and integrated over the horizon. Due to the ellipsoidal set description, linear dynamics and quadratic storage function, the problem is described by \emph{linear matrix inequalities} (LMIs) at each fixed time instant. By formulating $N$ number of LMIs uniformly distributed over the time horizon, a semidefinite program is set up to minimize $\rho$. 

Let $\rho^*$ be the optimal optimal value found by the algorithm. Then, it is guaranteed that the hook will be in a ball of radius $\rho^*$ around the reference position at the time of grasping. 
Hence, if the radius of the hook is larger than $\rho^*$, then the success of the payload grasping is ensured. In Section~\ref{sec:simu}, we show that this condition is satisfied in case of realistic uncertainty and disturbance magnitudes.
    
    

\begin{figure}
\vspace{2mm}
    \centering
    \ffigbox{\caption{Payload grasping from a moving platform on rough terrain.    \label{fig:terrain}}}{\includegraphics[width=\linewidth]{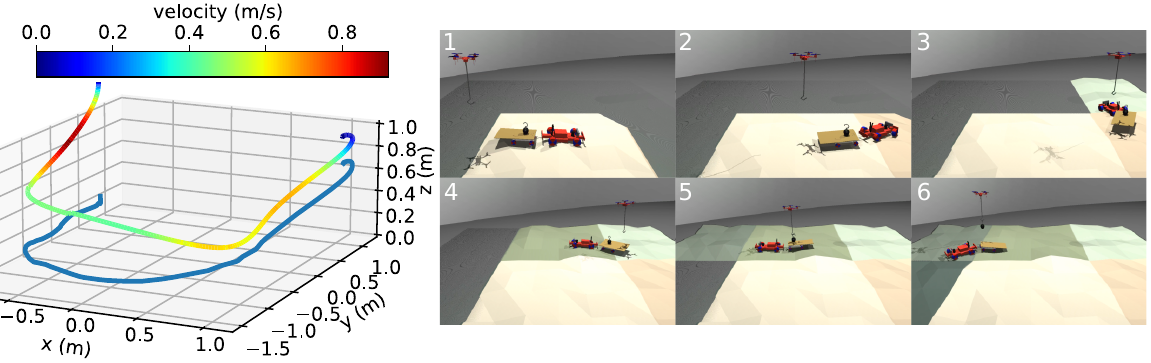}}
    \vspace{-7mm}
\end{figure}

\vspace{-1.5mm}
\section{Experiments}\label{sec:simu}

\begin{figure*}
\vspace{2mm}
    \centering \ffigbox{\caption{Motion trajectories of the hook of the quadcopter and the payload in real flight experiments with photos of the grasping time instants.
    \label{fig:meas}}}{\includegraphics[width=.95\linewidth]{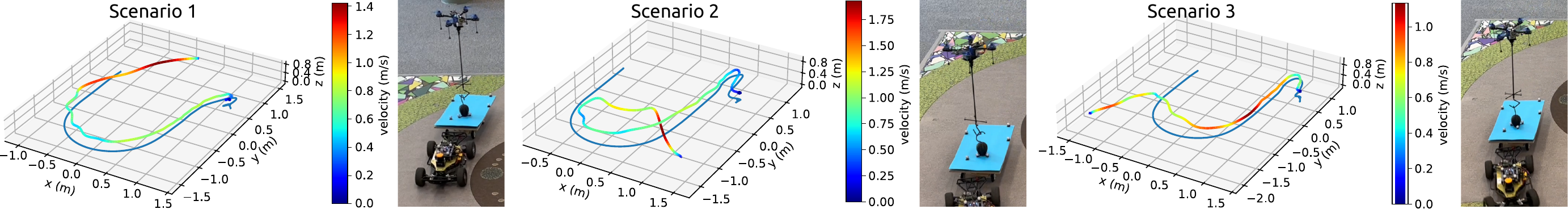}}
    \vspace{-6mm}
\end{figure*}

\vspace{-1.5mm}
\subsection{Experimental setup}
\vspace{-1.5mm}

In the experiments, we use an F1Tenth autonomous racecar \cite{OKelly20_f1tenth} as a UGV with the reference tracking LPV controller introduced in \cite{floch_gaussian_2024}. To carry the 3D printed payload, a custom designed trailer is attached to the UGV. The aerial platform is also custom designed (see \cite{antal2024autonomous} for details), equipped with a 3D printed hook connected to a carbon fiber rod. The experimental setup is shown in Fig.~\ref{fig:intro}, and the physical parameters of the manipulator are given in Table~\ref{tab:drone_param}.

For trajectory optimization, we use the following parameters: $h=0.05$ s, $T_\mathrm{d}=10$, $W_\mathrm{R}^i = 20 I_4$, $W^i_\mathrm{Q} = \mathrm{diag}(I_3, 0.1 I_3, 0_2, 3, 0.5 I_3, 0_2, I_2)$ for $i=1,2,3$. The weights of the LTV-LQR controller have been set to $\mathcal{W}_\mathrm{Q}=\mathcal{W}_\mathrm{F}=\mathrm{diag}(10I_3, I_2, 10, 0.1 I_3, I_3, 0.1 I_4)$, $\mathcal{W}_\mathrm{R}=\mathrm{diag}(5, 40 I_3)$. The numerical values of the parameters have been chosen based on preliminary simulation experiments.

\vspace{-1.5mm}
\subsection{Simulation study}
\vspace{-1mm}



To evaluate the performance of the proposed method, we use a paperclip shaped reference path for the UGV with constant speed, illustrated in Fig.~\ref{fig:paperclip}. Different scenarios are generated by changing the speed, the time window for grasping, and the initial pose of the quadcopter.

\textbf{Robustness analysis:} We perform the robustness analysis with an initial ellipsoid defined by $A_0=\mathrm{diag}(10 I_3, 100 I_3, 10 I_3, 100 I_3, 10 I_4)$, disturbance magnitude $\beta=0.02$, 
and $N=50$ discretization time steps. Moreover, as each reference trajectory results in a different LTV system, we solve the analysis problem for various designed trajectories using 16 initial drone positions (shown in Fig.~\ref{fig:paperclip}, with 1.1m initial height) and 3  UGV speed values (0.6, 0.8, 1.0 m/s), resulting in 48 optimizations altogether. Over the analyzed cases, the maximum error gain is $\rho^*=0.968$cm. The radius of the hook we use for the experiments is $r_\mathrm{hook}=2.3$cm, which is significantly larger than $\rho^*$, therefore we theoretically guarantee that the grasping is accomplished despite the considered uncertainties and disturbances.

\textbf{Payload grasping on rough terrain:} First, we analyze the performance of the proposed method by simulating a rough terrain that is discovered continuously, motivated by the problem investigated in \cite{aoki_hardware---loop_2023}. Each time a new part of the terrain is discovered, i) the trajectory of the UGV is simulated to update the payload motion prediction, ii) Optimization~\eqref{eq:planning_ocp} is solved to update the quadcopter reference trajectory, and iii) the LQR gains are recomputed and passed to the controller. The computation time of each step significantly depends on the duration of the maneuver, because it determines the number of simulation steps and SQP shooting nodes. For a maximum of $T_\mathrm{f}=17$s, the worst-case computation time of steps i)-iii) is 1.4s on a standard laptop. Hence, new trajectories are activated 1.4s after receiving the terrain information. By parallelizing the computations, the main simulation runs in real-time.

A specific scenario is illustrated in Fig.~\ref{fig:terrain}, where the upper plot shows the trajectory of the hook on the quadcopter with the path of the payload, and the lower plot shows snapshots from the MuJoCo simulation. The terrain includes 6 parts, each constructed by assigning random height values in the range of $[0, 0.12]$m to every node on a uniform grid with $0.25$m spacing, and connecting the nodes by flat surfaces. Subsequent terrain parts appear with at least 1.4s time difference, enabling real-time replanning of the reference trajectory. In this scenario, the mass of the payload is $m_\mathrm{load}=0.1$kg, the reference speed of the UGV is $0.5$m/s, the time window for grasping is specified as $\underline{T}_\mathrm{g}=7.5$s, $\overline{T}_\mathrm{g}=10.5$s, and the duration is $T_\mathrm{f}=16.5$s. The upper plot of Fig.~\ref{fig:terrain} shows that the manipulator follows the payload before grasping, which naturally results from the cost definition of Phase 1. Then, driven by the complementarity constraints, the quadcopter accelerates to grasp the payload. Finally, the payload is transported towards the target location. 

\vspace{-1mm}
\subsection{Real-world experiments}
\vspace{-1mm}
We evaluate the real-world performance of the proposed methods by flight experiments. Our quadcopter is equipped with a Crazyflie Bolt flight controller that runs the state estimation and LTV-LQR controller at 500Hz. Optitrack motion capture system is used to provide high-precision pose information of the vehicles and the payload. The computational budget of the microcontroller is not sufficient for nonlinear optimization, therefore the trajectories are computed on a desktop PC and sent to the drone via radio. However, our real-time simulations validate that a more powerful embedded computer would make it possible to run the whole algorithm onboard. \changes{Similarly, on-board (and possibly less accurate) sensors, such as camera, IMU, or GPS with suitable sensor fusion and state estimation could be applied to enable autonomous perception.}

Real-world experiments begin with collecting the pose of the manipulator, the UGV, the trailer, and the payload from the motion capture system. The UGV reference tracking is forward simulated, given the current position and velocity of the vehicle, to obtain the payload motion prediction, and the reference trajectory of the quadcopter is designed. Fig.~\ref{fig:intro} shows photos of a grasping experiment and the supplementary material contains videos of several scenarios\footnote{To put down the payload and land the quadcopter (shown in the supplementary video), we use the planning method of \cite{antal2024autonomous} with predefined waypoints, as these maneuvers are independent of the moving platform.}.

We investigate three scenarios in detail, which are shown in Fig.~\ref{fig:meas}, displaying the trajectory of the hook on the quadcopter, the position of the payload, and photos taken at the grasping time instants. These experiments have been conducted with payload mass $m_\mathrm{load}=0.1$kg, UGV reference speeds $\{0.6, 0.8, 0.8\}$m/s, time windows $\underline{T}_\mathrm{g} = \{5.5, 4.5, 4.2\}$s, $\overline{T}_\mathrm{g} = \{8.5, 7.5, 7.2\}$s, and durations $T_\mathrm{f} = \{14.5, 13.5, 13.2\}$s, respectively. As it is shown in the photos, the initial payload orientation has been different in each scenario. In Scenario~1, the payload forward-facing hook of the payload makes grasping more challenging than in the other configurations. However, according to \eqref{eq:grasp_cond}, the designed trajectory ensures that the payload is grasped from the side, leading to a successful maneuver. In Scenario~2, the acceleration of the hook-gripper is large at Phase~1 of the trajectory, and the agile motion leads to relatively large hook angles $\alpha$ and $\beta$. This example shows that although the hook is not actuated, it can be controlled precisely without reducing the agility of the motion. Finally, Scenario~3 shows a similar experiment with different parameters showing reliable performance of the method despite the variations of the initial conditions and specific parameter settings.

\vspace{-1mm}
\section{Conclusion}\label{sec:conclusion}
\vspace{-1mm}

In this work, we propose a complete solution for hook-based aerial payload grasping from a moving platform. The digital twin-based payload motion prediction and computationally efficient trajectory optimization enable reliable reference generation and rapid replanning in dynamically changing environments. We guarantee successful grasping under uncertainties and disturbances by IQC-based robustness analysis of the closed-loop manipulator dynamics. The experimental results show that the proposed approach is robust and efficient, enabling the hook-based manipulator to successfully grasp a small payload from a moving platform.

\changes{Future work includes further analysis to assess scalability for larger payloads and multi-agent scenarios, as well as development of on-board sensing and computation to facilitate fully autonomous operation and outdoor deployment.}

\bibliography{reference}

\begin{thebibliography}{10}
\providecommand{\url}[1]{#1}
\csname url@samestyle\endcsname
\providecommand{\newblock}{\relax}
\providecommand{\bibinfo}[2]{#2}
\providecommand{\BIBentrySTDinterwordspacing}{\spaceskip=0pt\relax}
\providecommand{\BIBentryALTinterwordstretchfactor}{4}
\providecommand{\BIBentryALTinterwordspacing}{\spaceskip=\fontdimen2\font plus
\BIBentryALTinterwordstretchfactor\fontdimen3\font minus
  \fontdimen4\font\relax}
\providecommand{\BIBforeignlanguage}[2]{{%
\expandafter\ifx\csname l@#1\endcsname\relax
\typeout{** WARNING: IEEEtran.bst: No hyphenation pattern has been}%
\typeout{** loaded for the language `#1'. Using the pattern for}%
\typeout{** the default language instead.}%
\else
\language=\csname l@#1\endcsname
\fi
#2}}
\providecommand{\BIBdecl}{\relax}
\BIBdecl

\bibitem{Ruggiero2018}
F.~Ruggiero, V.~Lippiello, and A.~Ollero, ``Aerial manipulation: {A} literature
  review,'' \emph{IEEE Robotics and Automation Letters}, vol.~3, pp.
  1957--1964, 2018.

\bibitem{Ollero2022}
A.~Ollero, M.~Tognon, A.~Suarez, D.~Lee, and A.~Franchi, ``Past, present, and
  future of aerial robotic manipulators,'' \emph{IEEE Transactions on
  Robotics}, vol.~38, no.~1, pp. 626--645, 2022.

\bibitem{baraban_fruit_grasping_2021}
G.~Baraban, S.~Kothiyal, and M.~Kobilarov, ``Perception-based {UAV} fruit
  grasping using sub-task imitation learning,'' in \emph{Proc. of the Aerial
  Robotic Systems Physically Interacting with the Environment}, 2021, pp. 1--8.

\bibitem{Maghazei2020}
O.~Maghazei and T.~Netland, ``Drones in manufacturing: Exploring opportunities
  for research and practice,'' \emph{Journal of Manufacturing Technology
  Management}, vol.~31, no.~1, pp. 1237--1259, 2020.

\bibitem{spica_aerial_2012}
R.~Spica, A.~Franchi, G.~Oriolo, H.~H. Bulthoff, and P.~R. Giordano, ``Aerial
  grasping of a moving target with a quadrotor {UAV},'' in \emph{{IEEE}/{RSJ}
  {International} {Conference} on {Intelligent} {Robots} and {Systems}}, 2012,
  pp. 4985--4992.

\bibitem{zhang_grasp_2018}
G.~Zhang, Y.~He, B.~Dai, F.~Gu, L.~Yang, J.~Han, G.~Liu, and J.~Qi, ``Grasp a
  moving target from the air: System \& control of an aerial manipulator,'' in
  \emph{{IEEE} {International} {Conference} on {Robotics} and {Automation}},
  2018, pp. 1681--1687.

\bibitem{luo_time-optimal_2023}
W.~Luo, J.~Chen, H.~Ebel, and P.~Eberhard,
  ``\BIBforeignlanguage{en}{Time-{Optimal} {Handover} {Trajectory} {Planning}
  for {Aerial} {Manipulators} {Based} on {Discrete} {Mechanics} and
  {Complementarity} {Constraints}},'' \emph{\BIBforeignlanguage{en}{IEEE
  Transactions on Robotics}}, vol.~39, no.~6, pp. 4332--4349, 2023.

\bibitem{Hua2021}
H.~Hua, Y.~Fang, X.~Zhang, and C.~Qian, ``A time-optimal trajectory planning
  strategy for an aircraft with a suspended payload via optimization and
  learning approaches,'' \emph{IEEE Transactions on Control Systems
  Technology}, vol.~30, no.~6, pp. 2333--2343, 2022.

\bibitem{Li2021}
G.~Li, A.~Tunchez, and G.~Loianno, ``{PCMPC}: Perception-constrained model
  predictive control for quadrotors with suspended loads using a single camera
  and {IMU},'' in \emph{Proc. of the IEEE International Conference on Robotics
  and Automation}, 2021, pp. 2012--2018.

\bibitem{li_autotrans_2023}
H.~Li, H.~Wang, C.~Feng, F.~Gao, B.~Zhou, and S.~Shen,
  ``\BIBforeignlanguage{en}{{AutoTrans}: {A} {Complete} {Planning} and
  {Control} {Framework} for {Autonomous} {UAV} {Payload} {Transportation}},''
  \emph{\BIBforeignlanguage{en}{IEEE Robotics and Automation Letters}}, vol.~8,
  no.~10, pp. 6859--6866, 2023.

\bibitem{wang_impact-aware_2024}
H.~Wang, H.~Li, B.~Zhou, F.~Gao, and S.~Shen, ``Impact-{Aware} {Planning} and
  {Control} for {Aerial} {Robots} {With} {Suspended} {Payloads},'' \emph{IEEE
  Transactions on Robotics}, vol.~40, pp. 2478--2497, 2024.

\bibitem{Thomas2014}
J.~Thomas, G.~Loianno, J.~Polin, K.~Sreenath, and V.~Kumar, ``Toward autonomous
  avian-inspired grasping for micro aerial vehicles,'' \emph{Bioinspiration \&
  Biomimetics}, vol.~9, no.~2, 2014.

\bibitem{antal2024autonomous}
P.~Antal, T.~Péni, and R.~Tóth, ``Autonomous hook-based grasping and
  transportation with quadcopters,'' \emph{IEEE Transactions on Control Systems
  Technology}, pp. 1--11, 2025.

\bibitem{Sreenath2013}
K.~Sreenath, T.~Lee, and V.~Kumar, ``Geometric control and differential
  flatness of a quadrotor uav with a cable-suspended load,'' in \emph{Proc. of
  the IEEE Conference on Decision and Control}, 2013.

\bibitem{Crousaz2014}
C.~D. Crousaz, F.~Farshidian, and J.~Buchli, ``Aggressive optimal control for
  agile flight with a slung load,'' in \emph{Proc. of the IROS 2014 Workshop on
  Machine Learning in Planning and Control of Robot Motion}, 2014.

\bibitem{sreenath_trajectory_2013}
K.~Sreenath, N.~Michael, and V.~Kumar, ``Trajectory generation and control of a
  quadrotor with a cable-suspended load - {A} differentially-flat hybrid
  system,'' in \emph{Proc. of the {IEEE} {International} {Conference} on
  {Robotics} and {Automation}}, Karlsruhe, Germany, 2013, pp. 4888--4895.

\bibitem{Beard2012}
R.~W. Beard and T.~W. McLain, \emph{Small Unmanned Aircraft: Theory and
  Practice}.\hskip 1em plus 0.5em minus 0.4em\relax Princeton University Press,
  2012.

\bibitem{thomson_vibration_2018}
W.~T. Thomson, \emph{Theory of Vibration with Applications}.\hskip 1em plus
  0.5em minus 0.4em\relax CrC Press, 2018.

\bibitem{todorov2012mujoco}
E.~Todorov, T.~Erez, and Y.~Tassa, ``Mujoco: A physics engine for model-based
  control,'' in \emph{Proc. of the IEEE International Conference on Intelligent
  Robots and Systems}, 2012, pp. 5026--5033.

\bibitem{lowrey_physically-consistent_2014}
K.~Lowrey, S.~Kolev, Y.~Tassa, T.~Erez, and E.~Todorov, ``Physically-consistent
  sensor fusion in contact-rich behaviors,'' in \emph{Proc. of the {IEEE}/{RSJ}
  {International} {Conference} on {Intelligent} {Robots} and {Systems}}, 2014,
  pp. 1656--1662.

\bibitem{bock_multiple_1984}
H.~G. Bock and K.~J. Plitt, ``A {Multiple} {Shooting} {Algorithm} for {Direct}
  {Solution} of {Optimal} {Control} {Problems},'' in \emph{Proc. of the 9th
  {IFAC} {World} {Congress}: {A} {Bridge} {Between} {Control} {Science} and
  {Technology}}, 1984, pp. 1603--1608.

\bibitem{verschueren_acadosmodular_2022}
R.~Verschueren, G.~Frison, D.~Kouzoupis, J.~Frey, N.~V. Duijkeren, A.~Zanelli,
  B.~Novoselnik, T.~Albin, R.~Quirynen, and M.~Diehl, ``acados—a modular
  open-source framework for fast embedded optimal control,'' \emph{Mathematical
  Programming Computation}, vol.~14, no.~1, pp. 147--183, 2022.

\bibitem{greer_shrinking_2020}
W.~B. Greer and C.~Sultan, ``Shrinking {Horizon} {Model} {Predictive} {Control}
  {Method} for {Helicopter}–{Ship} {Touchdown},'' \emph{Journal of Guidance,
  Control, and Dynamics}, vol.~43, no.~5, pp. 884--900, 2020.

\bibitem{Kwakernaak}
H.~Kwakernaak and R.~Sivan, \emph{Linear optimal control systems}.\hskip 1em
  plus 0.5em minus 0.4em\relax New York: Wiley Interscience, 1972.

\bibitem{seiler_finite_2019}
P.~Seiler, R.~M. Moore, C.~Meissen, M.~Arcak, and A.~Packard, ``Finite horizon
  robustness analysis of {LTV} systems using integral quadratic constraints,''
  \emph{Automatica}, vol. 100, pp. 135--143, 2019.

\bibitem{torrente_data-driven_2021}
G.~Torrente, E.~Kaufmann, P.~Fohn, and D.~Scaramuzza, ``Data-{Driven} {MPC} for
  {Quadrotors},'' \emph{IEEE Robotics and Automation Letters}, vol.~6, no.~2,
  pp. 3769--3776, 2021.

\bibitem{megretski_system_1997}
A.~Megretski and A.~Rantzer, ``System analysis via integral quadratic
  constraints,'' \emph{IEEE Transactions on Automatic Control}, vol.~42, no.~6,
  pp. 819--830, 1997.

\bibitem{veenman_robust_2016}
J.~Veenman, C.~W. Scherer, and H.~Köroğlu, ``Robust stability and performance
  analysis based on integral quadratic constraints,'' \emph{European Journal of
  Control}, vol.~31, pp. 1--32, 2016.

\bibitem{moore_finite_2015}
R.~M. Moore, ``Finite horizon robustness analysis using integral quadratic
  constraints,'' Master's thesis, University of California, Berkeley, 2015.

\bibitem{OKelly20_f1tenth}
M.~O'Kelly, H.~Zheng, D.~Karthik, and R.~Mangharam, ``F1tenth: An open-source
  evaluation environment for continuous control and reinforcement learning,''
  in \emph{Proc. of the NeurIPS Competition and Demonstration Track}, vol. 123,
  2019, pp. 77--89.

\bibitem{floch_gaussian_2024}
K.~Floch, T.~Péni, and R.~Tóth, ``Gaussian-process-based adaptive trajectory
  tracking control for autonomous ground vehicles,'' in \emph{Proc. of the
  European Control Conference}, 2024, pp. 464--471.

\bibitem{aoki_hardware---loop_2023}
N.~Aoki and G.~Ishigami, ``Hardware-in-the-loop {Simulation} for {Real}-time
  {Autonomous} {Tracking} and {Landing} of an {Unmanned} {Aerial} {Vehicle},''
  in \emph{Proc. of the {IEEE}/{SICE} {International} {Symposium} on {System}
  {Integration}}, 2023, pp. 1--6.

\end{thebibliography}

\end{document}